\documentclass[times,twocolumn,final]{elsarticle}

\usepackage{medima}
\usepackage{framed,multirow}

\usepackage{amssymb}
\usepackage{latexsym}

\usepackage{url}
\usepackage{xcolor}

\usepackage{hyperref}
\usepackage{lineno,hyperref}
\usepackage{graphicx}
\usepackage{subfig}
\usepackage{bm}
\usepackage{multirow}
\usepackage{textcomp}
\usepackage{todonotes}
\usepackage{amsmath}
\usepackage{diagbox}

\usepackage[linesnumbered,ruled,noline,noend]{algorithm2e}
\usepackage{setspace}

\SetKwBlock{ClientExecute}{Extract latent variables in institution ($k, E)$ : }{}
\SetKwBlock{ClientExecuteGenerator}{Updating auxiliary generator in institution k: }{}

\SetKwBlock{ClientExecuteGenerative}{Updating generator in institution ($k, E_{top}, E_{bottom}, D, \mathbf{e}$) :}{}
\SetKwBlock{ClientExecuteClassifier}{Updating classifier in institution ($k, D, \mathbf{e}, \theta_C $) :}{}

\SetKwBlock{ServerExecute}{Server executes:}{}
\SetKwFor{Forpar}{for}{do in parallel}{}
\SetKwInput{Input}{Input}{}
\SetKwComment{Comment}{$\triangleright$\ }{}

\def\widthfive{0.193\linewidth}
\def\widthfive1{0.175\linewidth}



\definecolor{newcolor}{rgb}{.8,.349,.1}

\journal{Medical Image Analysis}

\begin{document}

\begin{frontmatter}

\title{Handling Data Heterogeneity with Generative Replay in Collaborative Learning for Medical Imaging}


\author[1]{Liangqiong Qu}
\author[1]{Niranjan Balachandar}
\author[1]{Miao Zhang }
\author[2]{Daniel Rubin \corref{cor1}}
\ead{dlrubin@stanford.edu}
\cortext[cor1]{Corresponding author}

\address[1]{Department of Biomedical Data Science at Stanford
University, Stanford, CA 94305, USA}
\address[2]{Department of Biomedical Data Science  and Department
of Radiology at Stanford
University, Stanford, CA 94305, USA.}

\begin{abstract}
Collaborative learning, which enables collaborative and decentralized training of deep neural networks at multiple institutions in a
privacy-preserving manner, is rapidly emerging as a valuable technique in healthcare applications. However, its distributed nature often leads to
significant heterogeneity in data distributions across institutions. 
In this paper, we present a novel generative replay strategy to address the challenge of data heterogeneity in collaborative learning methods.  Different from traditional methods that directly aggregating the model parameters, we leverage generative adversarial
learning to aggregate the knowledge from all the local institutions. Specifically, instead of directly training a model for task
performance, we develop a novel dual model architecture: a primary model learns the desired task, and an auxiliary ``generative replay model'' allows aggregating knowledge from the heterogenous clients. The auxiliary model is then broadcasted to the central sever, to regulate the training of primary model with an unbiased target distribution. 
Experimental results demonstrate the capability of the proposed method in handling heterogeneous data across institutions. On highly heterogeneous data partitions, our model achieves $\sim$4.88\% improvement in the prediction accuracy on a diabetic retinopathy classification dataset, and $\sim$49.8\% reduction of mean absolution value on a Bone Age prediction dataset, respectively, compared to the state-of-the art collaborative learning methods.
\end{abstract}

\begin{keyword}
Collaborative learning, federated learning, data heterogeneity, Generative Adversarial Networks (GAN), autoencoder
\end{keyword}

\end{frontmatter}

\section{Introduction}

Deep artificial neural networks (DNNs) have led to state-of-the-art performances in a wide range of computer vision tasks.
One major issue in DNNs is the requirement for large quantities of the annotated image data to train robust models. However, large quantities of the data are inherently decentralized and are widely distributed in multiple centers, especially for medical imaging, where cohort sizes at single institutions are often small.
Gathering data from multiple centers is often hindered by barriers to data sharing and regulatory and privacy concerns. Collaborative learning (also termed federated learning, distributed learning, or decentralized learning) allows collaboratively training a shared global model on multiple centers while keeping the personal data decentralized, and is thus an attractive alternative \citep{mcmahan2016communication,chang2018distributed,kairouz2019advances}.

Recently, collaborative learning has been highlighted as a foundational research area in medical domain \citep{langlotz2019roadmap}. Numerous collaborative learning methods have been proposed in past decades \citep{su2015experiments,mcmahan2016communication,lin2017deep,chang2018distributed,vepakomma2018split,qu2021rethinking}, such as Federated Averaging (FedAVG) \citep{mcmahan2016communication} and Cyclical weight transfer (CWT) \citep{chang2018distributed}. Despite the promising progress, collaborative learning still faces the data heterogeneity challenge, where data across institutions is usually non-independent and identically distributed (non-iid)~\citep{qu2021experimental,roth2020federated,dayan2021federated,liu2021feddg}, which impedes the model convergence and cause performance drops.
Numerous efforts have been devoted to solving the performance drops on data heterogeneity challenge \citep{hsu2019measuring,hsieh2019non,Niranjan2020}, such as applying different types of optimization heuristic to stabilize the local model update \citep{li2020federated,hsu2019measuring, Hsu2020ECCV,wang2020federated}, and using knowledge distillation to learn a more powerful global model \citep{Gong_2021_ICCV,Zhang_2021_ICCV}. 
While effective, these methods may require heuristic parameters tune \citep{hsu2019measuring,hsieh2019non,li2020federated}, suffer from usage restriction for layer-wise mapping optimization methods~\citep{Zhang_2021_ICCV}, or may not be optimal for highly heterogeneous settings (see Fig.~\ref{fig:performance}).

\begin{figure*}[t]
	\centering
	\includegraphics[width=0.95\linewidth]{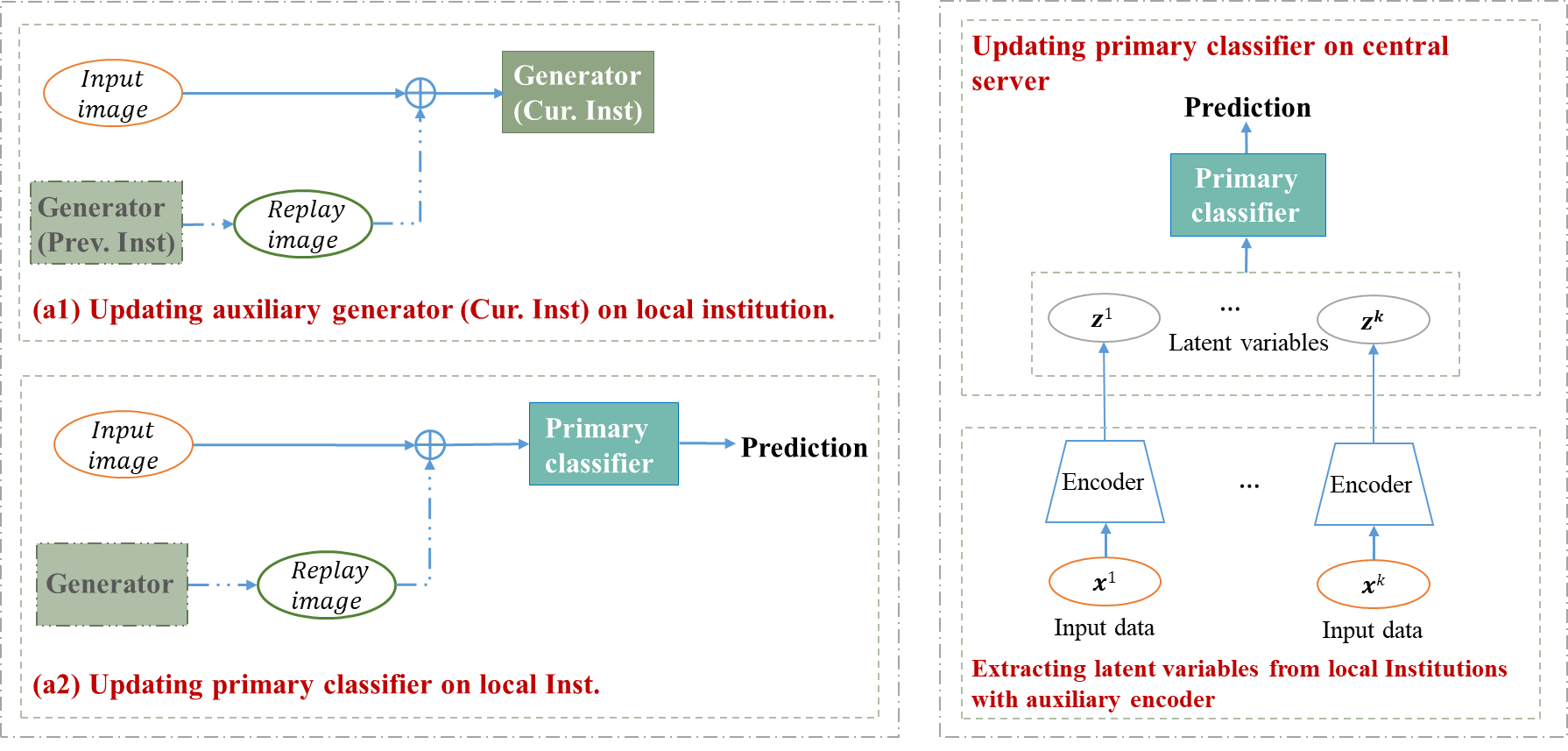}
	\vspace{-5.0mm}
	\caption[caption]
{Two ways of applying the proposed generative replay strategy to address the challenge of data heterogeneity among institutions that participate in collaborative learning. We apply an image classification task as the desired task.  Left: Illustration of incorporating the proposed generative replay strategy into existing collaborative method CWT~\citep{chang2018distributed} (referring to CWT+Replay). Right: Illustration of applying the proposed generative replay strategy to construct an individual collaborative learning framework FedReplay.}
	\label{fig:framework_union}
	\vspace{-2mm}
\end{figure*}

In this paper, we propose a novel and flexible generative replay strategy to address the data heterogeneity challenge in collaborative learning and to meet gaps in the prior approaches.
Different from traditional methods that directly aggregating the model parameters, we leverage generative adversarial learning \citep{shin2017continual,van2018generative,choi2018stargan} to aggregate the knowledge from the data distribution of all the local institutions. 
Specifically, we develop a novel dual model architecture consisting of a primary classifier for the desired task (e.g., a classification or regression task), and an auxiliary ``generative replay model'' for either generating artificial examples or extracting latent features from local clients. Our generative replay strategy helps solve the challenge of data heterogeneity in collaborative learning by training the primary model using the aggregated knowledge collected by generative replay model from the heterogenous clients. 

Our strategy is flexible to deploy,
can either be 1) incorporated into a well-deployed collaborative learning methods to improve their capability of handling data heterogeneity across institutions (left image in Fig.~\ref{fig:framework_union}), or 2) be used to construct a novel
and individual collaborative learning framework FedReplay (right image in Fig.~\ref{fig:framework_union}) for communication efficiency and better privacy protection. When incorporated into an existing collaborative learning framework---for example CWT \citep{chang2018distributed}---we first train an auxiliary and universal image generation network on local institutional data, share the generator between local institutions, and then train a standard CWT model based on the local institutional data and the augmented replayed data from the shared generator (referred to CWT+Replay). The training of auxiliary generator in CWT+Replay is time-consuming and the replayed data may reveal patient privacy.
We further apply our generative replay technique to conduct an individual collaborative learning framework, termed FedReplay, to avoid the possibility of revealing patient privacy from the replayed images.  Similar to CWT+Replay, FedReplay also has a dual model
architecture, a primary model that learns the desired task, and an auxiliary encoder network that extracts latent variables.
Our FedReplay is robust to various types of heterogeneity in data across institutions, as the primary model is directly trained on the union of the latent variables collected from the auxiliary generators of all the local institutions, and the latent variables contain most important features in the original data. The main contributions of our paper are summarized as follows:

1) We propose a novel generative replay technique to address the challenge of data heterogeneity among institutions that participate in collaborative learning. As opposed to existing methods that either provide sophisticated ways
to control the optimization strategy or share partial global data
to mitigate the performance drops from data heterogeneity, our generative replay technique provides a new insight for the development of collaborative learning methods, which is easy to implement, and can be applied to any kind of deep learning task (e.g., classification, regression, etc.).

2) Our generative replay technique is flexible to use. It can either be directly incorporated into a existing federated learning framework to increase their capability of handling data heterogeneity across institutions with minimal modifications (left image in Fig.~\ref{fig:framework_union}), or be used as a novel and individual collaborative learning framework to reduce communication cost and mitigate privacy cost (right image in Fig.~\ref{fig:framework_union}).

3) While previous collaborative learning methods require frequent communication between local institutions and the central server, the proposed FedReplay only requires a one-time communication between local institutions and the central server, which is time-efficient. The training of primary model is performed solely on the central server, without restriction to hardware
and network speeds among institutions.

\section{Related Works }
\textbf{Collaborative Learning} \quad
Prior collaborative learning methods can be generally classified into two categories, parallelized collaborative learning methods and serial collaborative learning methods. 

Parallelized methods involve training each individual institution on the local institutional data for several iterations/epochs, transferring the gradients/weights from individual institutions to a central server for averaging, and then transferring the averaged weights/gradients back to individual institutions. FedAVG \citep{mcmahan2016communication} and Federated stochastic gradient descent (FedSGD) \citep{su2015experiments,mcmahan2016communication} are two of the most popular parallelized methods. FedAVG \citep{mcmahan2016communication} involves frequent transferring of model weights between individual institutions and central server, while FedSGD \citep{su2015experiments,mcmahan2016communication} involves frequent transferring of gradients between individual institutions and central server, which is an extreme case of FedAVG \citep{mcmahan2016communication}.

Parallelized methods are computationally efficient due to their parallel processing. However, they may be
suboptimal if institutions have very different network connection speeds or deep learning hardware (a common
situation among medical institutions). \citet{bonawitz2019towards} relaxed the strict system condition by dropping devices that fail to compute the pre-defined epochs within a specified time window.  Beyond the heterogeneity in system, heterogeneity in data across institutions is another critical concern for parallelized methods. The training examples at each local institution are sampled from institution-specific distribution for parallelized methods, which is a biased estimator of the central target distribution if heterogeneity exists. The learned weights in different institutions will diverge severely when high skewness exists in the data, thus the synchronized averaged central model will lose accuracy or even completely diverge \citep{hsu2019measuring}. Several recent efforts have been devoted to solving the performance drops in statistically data heterogeneous settings. For example, \citet{zhao2018federated} proposed FedAVG+Share to improve the performance of FedAVG on non-IID data by distributing a small amount of data for globally sharing between all the institutions. Unlike FedAVG that simply updates the weights on the server, \citet{hsu2019measuring} accumulated the model updates with momentum and used an exponentially weighted moving average as the model update, to improve its robustness to non-IID data. While effective, these methods either require 1) all the participated institutions are active \citep{khaled2019first,wang2018cooperative}, 2) additional convexity assumptions \citep{wang2019adaptive}, 3) need sharing partial institutional data \citep{zhao2018federated}, or 4) only work well in data distributions with mild heterogeneity (see the comparison results in Fig.~\ref{fig:performance}).

Compared to the parallelized methods, serial methods involve training in a serial and cyclical way, which may be less computationally efficient than parallel processing, but may be more flexible to variations in hardware
and network speeds among institutions. CWT \citep{chang2018distributed} is one of the typical serial methods, which involves updating weights at one institution at a time, and cyclically transferring weights to the next training institution until convergence.
Split learning (SplitNN) \citep{vepakomma2018split} can be also considered as a serial collaborative learning method. In SplitNN \citep{vepakomma2018split}, each institution trains a partial deep network up to the cut layer, sends the intermediate output feature maps at the cut layer to the server. The server then completes the rest
of the training without looking at raw data from clients. In SplitNN, in addition to the model weights, the intermediate feature maps and gradients are communicated between different institutions. Similar to parallelized methods, heterogeneity in data across institutions is also a big concern in serial methods.
Serial methods always suffer from catastrophic forgetting problems when heterogeneity exists in data distributions. The model tends to abruptly forget what was previously learned information when it transfers to the next institution. The higher the skewness of heterogeneity in data, the more severe the catastrophic forgetting tends to be in serial methods, thus resulting in performance drops. \citet{Niranjan2020} introduced cyclically weighted loss to mitigate the performance loss for label distribution skewness in CWT, but this classification label based weighted loss only works for image classification tasks. By contrast, our generative replay strategy is more scalable and can work well on various types of tasks.

\textbf{Adversarial Attacks} \quad Recent works have shown that collaborative learning is vulnerable to gradient inversion attacks \citep{zhu2020deep,huang2021evaluating,geiping2020inverting} or model inversion attacks \citep{yin2021see,fredrikson2015model,he2019model}. \citet{zhu2020deep} demonstrated that it was able to reconstruct a client's private data with the shared gradients. But this work is limited to shallow network and low resolution images. \citet{geiping2020inverting} substantially improved the reconstructed image quality by exploiting a cosine distance loss together with the optimization problem. Similar to gradient inversion attacks, model inversion attacks are first introduced by \citet{fredrikson2015model} and aim to reconstructing private data from the output (or intermediate output) through the inference of a well-trained regression model. Recent works have improved and extended the approach to more complex setting \citep{yin2021see,fredrikson2015model,he2019model}, e.g., extending to modern DNNs \citep{he2019model} on collaborative learning. In this paper, we study the privacy protection capability of the different collaborative learning methods under gradient inversion attacks and model inversion attacks.

\begin{algorithm*}[t]
\renewcommand{\thealgocf}{1}
	\caption{CWT+Replay. The $K$ institutions are indexed by $k$, and $\mathbf{x}_k$ is training data for institution $k$. $N$ and $M$ are the training epochs, $\theta_C$ and $\theta_G$ are the model weights, $L$ and $L_G$ are the loss functions, and $\eta$ and $\eta_G$ are the learning rate for primary classifier and auxiliary generator, respectively. $E_{top}$ and  $E_{bottom}$ are encoder functions, ${D}$ is the decoder function, and $\textbf{e}$ is the encoded latent variables.}
	\label{algo:CWT_replay_algorithm}
\ClientExecuteClassifier{	
\For{\upshape{local epoch $i$ in epochs $N$}}{
\For{\upshape{each local batch} $x_k$ and $e$ \upshape{in} ${\mathbf{x}_k}$ and $\mathbf{e}$}
{$\widetilde {{x^r}}$ $\gets$$D(e)$ \quad  \Comment{Synthesizing replayed images}
$\mathbf{x_U}$ $\gets$ \{$\widetilde {x^r}, x_k$\}  \   \Comment{Union of synthetic and real images}
$\theta_{C} \gets \theta_{C} - \eta \nabla L(\theta_C; \mathbf{x_U})$ \quad \Comment{Updating classifier}
        }}
return $\theta_{C}$ to server
}

\ClientExecuteGenerative{
$D_{old} \gets D$ \\
\For{\upshape{local epoch $i$ in epochs $M$}}{
\For{\upshape{each local batch} $x_k$ and $e$ \upshape{in} ${\mathbf{x}_k}$ and $\mathbf{e}$}
{
$\widetilde {{x^r}}$ $\gets D_{old}(e)$ \quad \Comment{Synthesizing replayed images}

$x_U$ $\gets$ \{$\widetilde {x^r}, x_k$\} \  \Comment{Union of synthetic and real images}

$\mathbf{h}_{top} \gets E_{top}(x_U)$

$\mathbf{e}_{top} \gets Quantize(\mathbf{h}_{top}) $ \quad \Comment{Quantize}

$\mathbf{h}_{bottom} \gets E_{bottom}(x_k, \mathbf{e}_{top})$

$\mathbf{e}_{bottom} \gets Quantize(\mathbf{h}_{bottom}) $  \quad \Comment{Quantize}

$\widetilde {{x_U}}$ $\gets$ D($\mathbf{e}_{top}, \mathbf{e}_{bottom})$

$\theta_{G} \gets \theta_{G} - \eta_G \nabla L_G(\theta_G;  x_U, \widetilde {{x_U}})$ \quad \Comment{Updating generator} 

$\mathbf{e}$ $\gets$ \{$ \mathbf{e}_{bottom} , \mathbf{e}_{top}$\}

}
return $E_{top}, E_{bottom}, D, \mathbf{e}$  to server
}
}

\ServerExecute{
Initialize parameters $\theta_C$ and parameters $\theta_G$

\For{\upshape{each round} $t = 1, 2, ... $  }{
\For{\upshape{each institution in}  $k = 1, 2, ..., K $}
{
\eIf {$t == 1$} {
    $(E_{top} ,E_{bottom} ,D,\mathbf{e}) $ $\gets$ \textbf{Updating generator in institution $(k, E_{top} ,E_{bottom} ,D, \mathbf{e})$}
}{$(D,\mathbf{e}, \theta_C) $ $\gets$ \textbf{Updating classifier in institution $(k, D,\mathbf{e}, \theta_C)$}
}
}}

}
\end{algorithm*}

\section{Method}
In this section, we 1) first describe the details of how to incorporate the proposed generative replay strategy into existing collaborative learning networks, 2) present the details of the proposed individual collaborative method FedReplay, 3) and provide the privacy analysis under different adversarial attacks. 

\subsection{Optimizing Collaborative Learning Methods with Generative Replay}
In this section, we use CWT \citep{chang2018distributed} as an example to illustrate how to apply the proposed generative replay strategy to the existing collaborative learning methods. We use an image classification task as the desired task. We apply the Vector Quantized Variational AutoEncoder (VQ-VAE-2) \citep{razavi2019generating} as our auxiliary image generation network.
 The complete pseudo-code of CWT+Replay is shown in Algorithm~\ref{algo:CWT_replay_algorithm} and its detailed architecture is shown in Fig.~\ref{fig:framework_union}.

CWT+Replay consists of a primary model (classifier) for the learned classification task, and an auxiliary generative replay model for synthesizing images that closely resemble input images. In CWT+replay, the auxiliary generator (VQ-VAE-2 \citep{razavi2019generating}) is first trained before the standard CWT training of primal model. VQ-VAE-2 is originated from Vector Quantized Variational AutoEncoder (VQ-VAE) \citep{van2017neural}, which consists of an encoder for mapping the observations to discrete latent variables, and a decoder for reconstructing observations from these latent variables, and its prior is learnt rather than static. Similar to CWT \citep{chang2018distributed}, the generator is also trained in a serial way, i.e., first updating
generator weights $\theta_G$ at one institution at a time, and serially transferring weights to the next training institution. During the training of generator in institution $k$, the generator aims to learn a mixed data distribution of both the real data $\mathbf{x}_k$ in institution $k$ and the replayed data from previous generator. Here total $K$ institutions are involved in the collaborative learning framework. 

Once the generator is trained successfully, we then distribute the decoders, the latent variables\footnote{We can also train a prior generator with PixelCNN \citep{oord2016pixel} to simulate the latent variables, and then distribute the trained PixelCNN instead of the latent variables.}, and their corresponding labels to all the institutions for standard primary classifier training.
Specifically, during the training of the classifier in institution $k$, the classifier is trained based on both the real data $\mathbf{x}_k$ in institution $k$ and the replayed data from the generator. With the auxiliary replayed data, all the institutions now aim at optimizing the primary classifier based on both the replayed data which resemble the data distribution from other institutions and the institutional specific data, rather than only the institution-specific data distribution, thus do not suffer from catastrophic forgetting as in CWT \citep{chang2018distributed}.

\begin{algorithm}[h]
\renewcommand{\thealgocf}{2}
	\caption{FedReplay. The $K$ institutions are indexed by $k$. $\mathbf{x}_k$ is training data for institution $k$. Model weights $\theta_C$, loss function $L$, learning rate $\eta$, and total training epochs $N$ for primary classifier, respectively. $E$ is the auxiliary pre-trained encoder function.}
	\label{algo:fedreply}
\ClientExecute{	
\Comment{Extract latent variables}
$\mathbf{e}_{k} \gets E(\mathbf{x}_k) $

\Comment{Return latent variables to server}
return $\mathbf{e}_{k}$ to server
}

\vspace*{0.1in}
\ServerExecute{
\Comment{Extracting the latent variables from all the institutions}
\Forpar{\upshape{each institution in}  $k = 1, 2, ..., K $}{
${\mathbf{e}_k}$  $\gets$ \textbf{Extract latent variables in institution $(k,E_{top} ,E_{bottom}$ )}	

}

\Comment{Union of all latent variables}

$\mathbf{e}$ $\gets$ \{$ {\mathbf{e}_1},  {\mathbf{e}_2}, ...,  {\mathbf{e}_K}$\}

\vspace*{0.035in}
\Comment{Updating primary classifier in server with latent variables}
Initialize parameters $\theta_C$

\For{\upshape{each batch $e$ in $\mathbf{ e}$ in total epochs $N$} }{
$\theta_{C} \gets \theta_{C} - \eta \nabla L(\theta_C; e)$
}
}
\end{algorithm}

The application of generative replay strategy to other collaborative learning methods can be also deployed similarly. For example, when incorporating into FedAVG \citep{mcmahan2016communication}, similar to CWT+Replay, we first train an auxiliary generative replay model to synthesize images that closely resemble the input images, we then train a standard parallelized FedAVG model based on both the real data from local institutions and the replayed data from the generator.

\subsection{The FedReplay Collaborative Learning Approach}
Even though CWT+Replay works well on non-IID situations, it still requires frequent transfer of the model weights between local institutions and the central server. In addition, the training of auxiliary image generation network for CWT+Replay is time-consuming and the replayed data may reveal patient privacy. To optimize the communication cost and to prevent privacy leakage from replayed data, we introduce a new collaborative learning approach, FedReplay, based on our generative replay technique in this section. Similar to CWT+Replay, FedReplay also uses a dual model architecture, where an auxiliary encoder extracts the latent variables from local institutions, and a primary model learns the task based on the extracted latent variables.

The framework of our FedReplay is shown in Fig.~\ref{fig:framework_union} and its pseudo-code is depicted in Algorithm \ref{algo:fedreply}.
FedReplay assumes the server has high storage capacity, which allows storage of the relevant latent variables, and training of the primary model based on the extracted latent variables. Our FedReply follows a three-stage training phase: 1) Train a universal and auxiliary encoder network on the data at one of the local institutions and distribute the unique pre-trained encoder network to all of the other institutions; 2) Apply the pre-trained encoder network to local institutions, compress the raw data into latent variables, and upload the latent variables and their corresponding labels to the central server; 3) Train a global primary model purely on the central server with the union of all the latent variables. 
As we do not need to know the decoder part for data reconstruction in FedReplay, we can simply apply several convolutional and ReLu (the nonlinear rectified linear unit function) layers as the encoder network. In addition, the training of the auxiliary encoder network can be also easily achieved on one
of the arbitrary institutions with the learned task (see Section \ref{section:experimental Setup} and Fig.~\ref{fig:FedReplay_network} for more details).

Our FedReplay is robust to various types of heterogeneity in data across institutions, as the primary model is directly trained on the union of the latent variables from all the institutions, and the latent variables contain most important features in the original data. FedReplay is also robust to the system heterogeneity, since the training of primary model is totally performed on the central sever, and it only requires a one-time communication between institutions and the central server. 

\subsection{Adversarial Attacks}
Recent works have shown that it is able to reconstruct pixel-wise level private data from the shared gradients or latent variables in collaborative learning~\citep{he2019model,zhu2020deep}.
We evaluate the privacy protection capability of FedReplay by comparing the quality of recovered image in FedReplay and the baseline FedAVG under model inversion attacks ~\citep{he2019model} and gradient inversion attacks ~\citep{geiping2020inverting}, respectively.

FedAVG shares the gradients of the whole network, while FedReplay shares the intermediate latent variables extracted from the auxiliary encoder network. Given a neural network with parameters $\theta$, the real private image $x$ that we aim to recover, the shared gradient $\nabla L_\theta(x)$, and the dummy input $\tilde x$, we formulate the image recovery task of FedAVG as an optimization problem following ~\citep{geiping2020inverting}:
\begin{equation}\label{FedAVG_DL}
  \arg \mathop {\min }\limits_{\tilde x} \left( {1 - \frac{{\left\langle {{\nabla _\theta }{L_\theta }(\tilde x),{\nabla _\theta }{L_\theta }(x)} \right\rangle }}{{\left\| {{\nabla _\theta }{L_\theta }(\tilde x)} \right\|\left\| {{\nabla _\theta }{L_\theta }(x)} \right\|}}} \right) + \alpha {\rm{TV}}(\tilde x).
\end{equation}
The former part aims to match the gradient of the recovered input $\tilde x$ with the target transmitted gradient ${\nabla _\theta }{L_\theta }(x)$, which is measured as a cosine distance following ~\citep{geiping2020inverting}. The latter part regularizes the recovered image with a total variation \citep{rudin1992nonlinear}, which encourages $\tilde x$ to be piece-wise smooth. The hyperparameter $\alpha$ balances the effects of the two terms.

Similarly, malicious attacker may also be able to recover the private data from the shared latent variables (denoted as ${E_\theta }(x)$) in FedReplay ~\citep{he2019model,zhang2020secret}. Specifically, we consider the white-box setting for this model inversion attack, as each local client shares the same auxiliary encoder network.
We formulate the model inversion attacks as the following optimization problem:
 \begin{equation}\label{FedReplay_DL}
  \arg \mathop {\min }\limits_{\tilde x} {\left\| {{E_{\theta }}(\tilde x) - {E_\theta }(x)} \right\|^2} + \alpha {\rm{TV(}}\tilde x{\rm{)}},
\end{equation}
where the former part is Euclidean Distance which aims to match the latent variables ${E_{\theta }}(\tilde x)$ of dummy input to the shared latent variables ${E_\theta }(x)$. Following ~\citep{he2019model}, we apply regularized Maximum Likelihood Estimation to solve this problem. We will qualitatively and quantitatively measure the privacy protection capability of FedAVG and FedReplay according to Eq.~\ref{FedAVG_DL} and Eq.~\ref{FedReplay_DL} in the \ref{section:exp_privacy}.

\section{Experiments}
We applied our generative replay approach to CWT \citep{chang2018distributed}, and compared the proposed CWT+Replay and FedReplay, with several state-of-the-art collaborative learning methods \citep{mcmahan2016communication,chang2018distributed,vepakomma2018split,hsu2019measuring,zhao2018federated,li2020federated}. In this section, we will describe the studied
dataset, detailed simulated data partitions, the compared methods, and quantitative results to demonstrate the effectiveness of the proposed generative replay strategy. A privacy analysis for the proposed generative replay strategy is also provided at the end of this section.

\subsection{Experimental Setup}\label{section:experimental Setup}

\textbf{Dataset.}
We evaluated the performance of different collaborative learning methods on both medical image classification and regression tasks.
\begin{itemize}
  \item \textbf{Image classification task on Diabetic Retinopathy (Retina) Dataset} \citep{KaggleRetina}. Retina dataset consists of 17563 pairs of right and left color digital retinal fundus images. Each image was rated as a scale of 0 to 4 according to the presence of diabetic retinopathy, where 0 to 4 indicates no, mild, moderate, severe and proliferative diabetic retinopathy, respectively. For simplicity, the labels were binarized to Healthy (scale 0) and Diseased (scale 2, 3 or 4) in our study, while the mild diabetic retinopathy images (scale 1) between healthy and diseased status were excluded. Additionally, only left eye images were used, to remove the confusion from using multiple images for the same patient. We randomly selected 6000 images (3000 positive (Healthy) and 3000 negative (Diseased) images) for training, 3000 images (1500 positive and 1500 negative images) for validation, and 3000 images (1500 positive and 1500 negative images) for testing. We applied a similar technique that Kaggle Diabetic Retinopathy Competition winner Benjamin Graham proposed \citep{graham2015kaggle} to pre-process these images, i.e., first rescaling images to the same eye radius of 300, subtracting the local average color of each image, then cropping to remove the image boundaries, and finally resizing to image resolution $256 \times 256$ to serve as the input of deep neural network. Similar to \citep{chang2018distributed}, the prediction accuracy was applied as the evaluation metric.

  \item \textbf{Image recognition task on RSNA Pediatric Bone Age Prediction with X-Ray image (BoneAge dataset)} \citep{halabi2019rsna}.  BoneAge dataset is used to estimate the bone age of pediatric patients based on radiographs of their hand. It consists of 14236 (mean age, 127 months) labeled and deidentified hand radiographs. Only male hands were used in our study, to remove the discrepancy between male and female hands. Specifically, we randomly selected 1000 images for validation, used 4575 images for training, and applied the original test dataset from BoneAge dataset as the global test dataset. The image resolution of the bone age images were resized to $256 \times 256$, and their intensity values were normalized to [0,1]. The mean absolute distance in months was used as the evaluation metric.
\end{itemize}
\begin{figure*}[t]
\scriptsize
	\begin{center}
		\begin{tabular}{ccc}
 \includegraphics[width=0.3\linewidth]{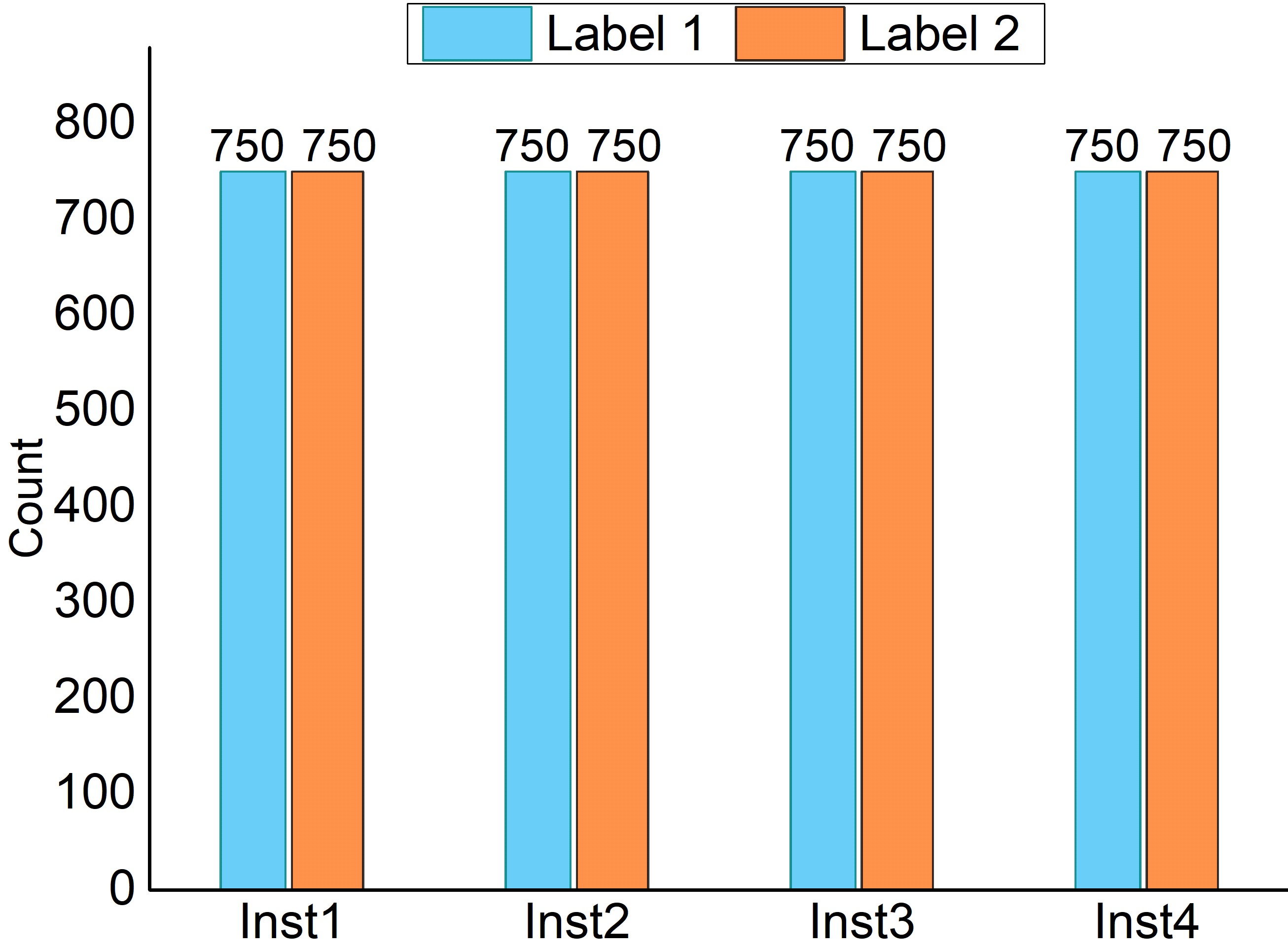}&
 \hspace{-3.7mm}
 \includegraphics[width=0.3\linewidth]{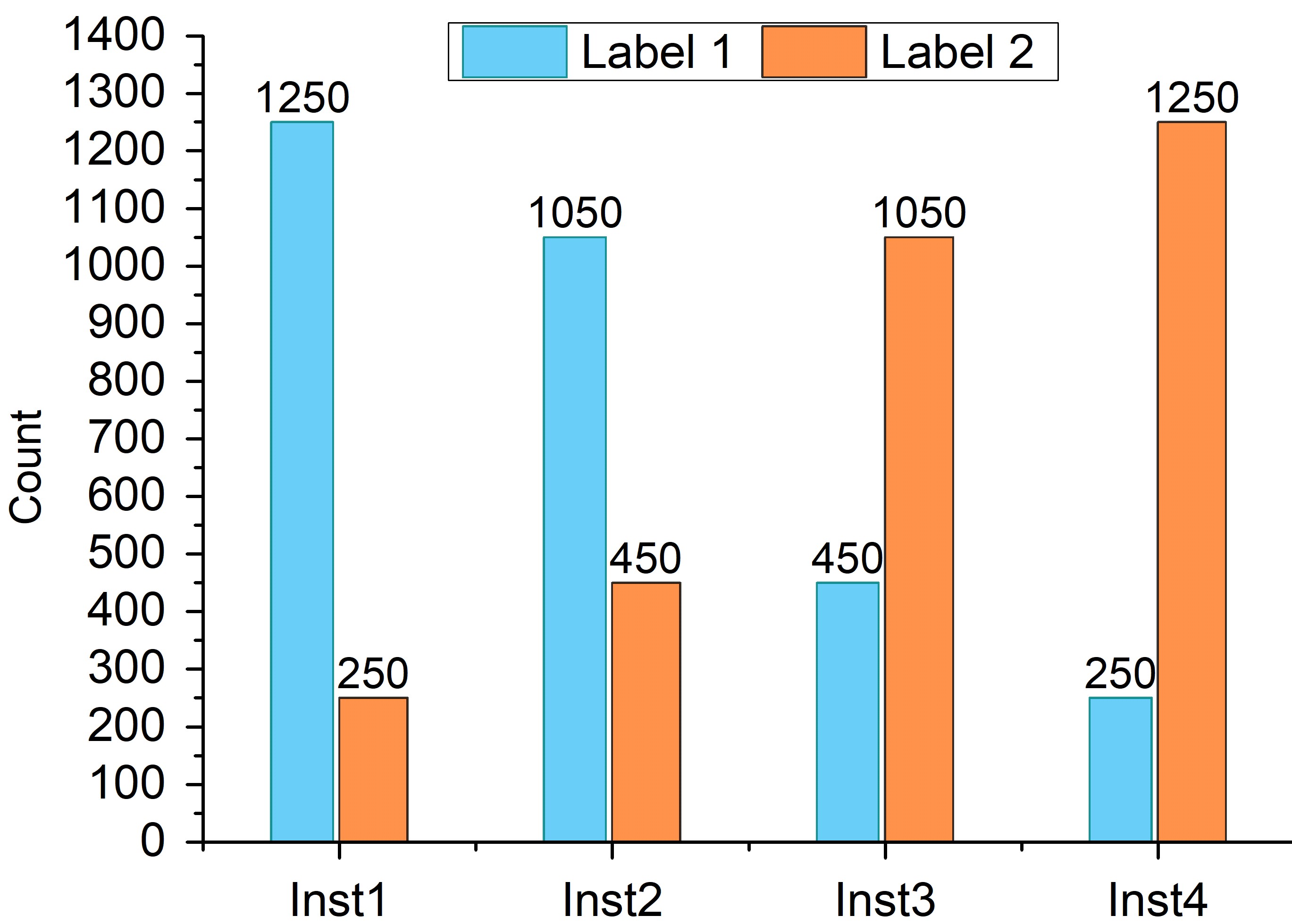} &
  \hspace{-3.7mm}
 \includegraphics[width=0.3\linewidth]{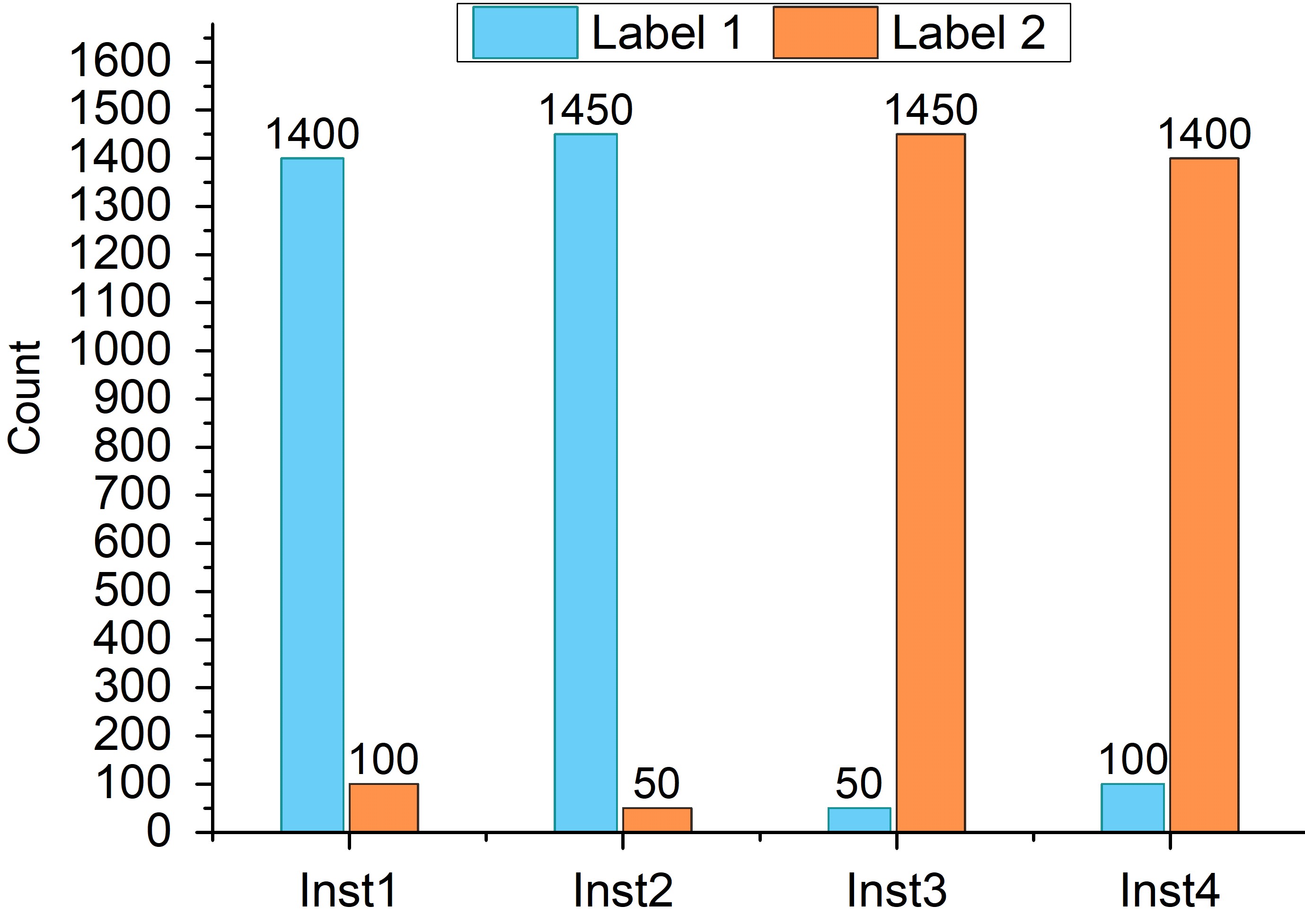} \\
 (a) Split 1, KS-0 & (b) Split 2, KS-0.40 & (c) Split 3, KS-0.61\\
		\end{tabular}
	\end{center}
	\vspace{-2mm}
	\caption{Three sets of simulated data partitions on Retina dataset \citep{KaggleRetina}. Here labels 1 and 2 indicate positive (Healthy) and negative (Diseased) labels, respectively. Large KS indicates high degree of label distribution skewness.}
	\label{fig:Retina_Split}
\vspace{-2mm}
\end{figure*}

\begin{figure*}[t]
\scriptsize
	\begin{center}
		\begin{tabular}{ccc}
 \includegraphics[width=0.3\linewidth]{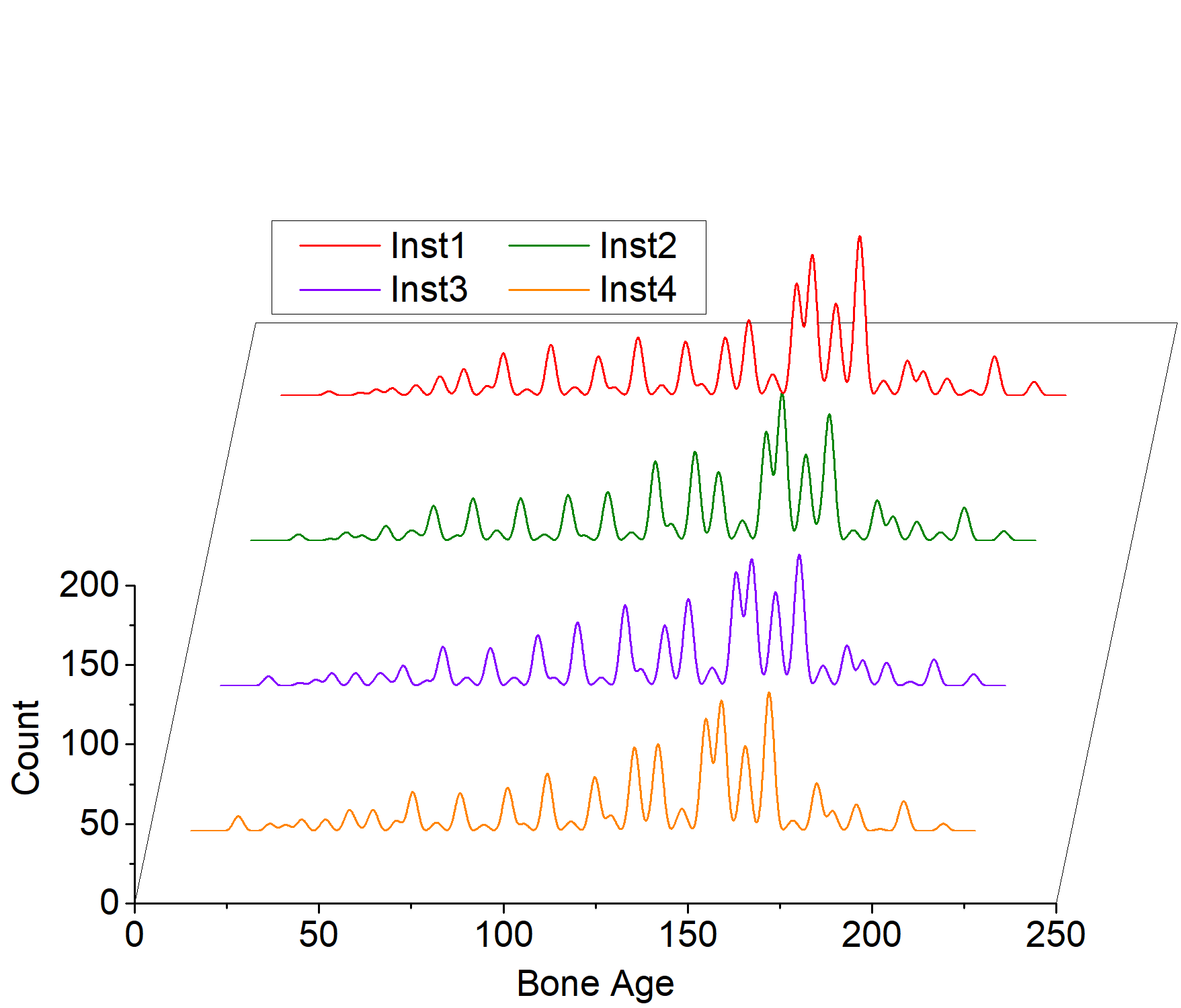} &
  \includegraphics[width=0.3\linewidth]{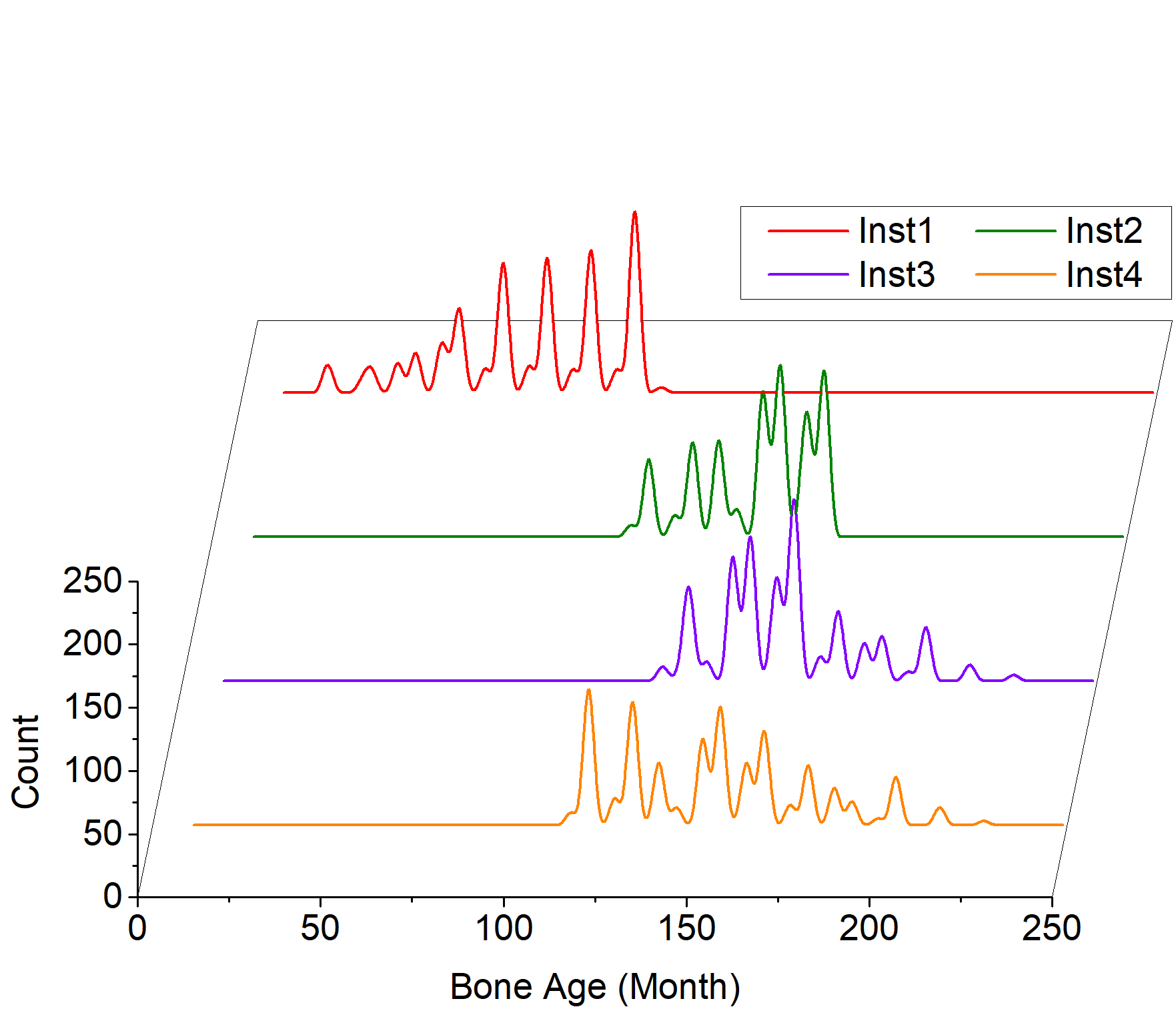} &
 \includegraphics[width=0.3\linewidth]{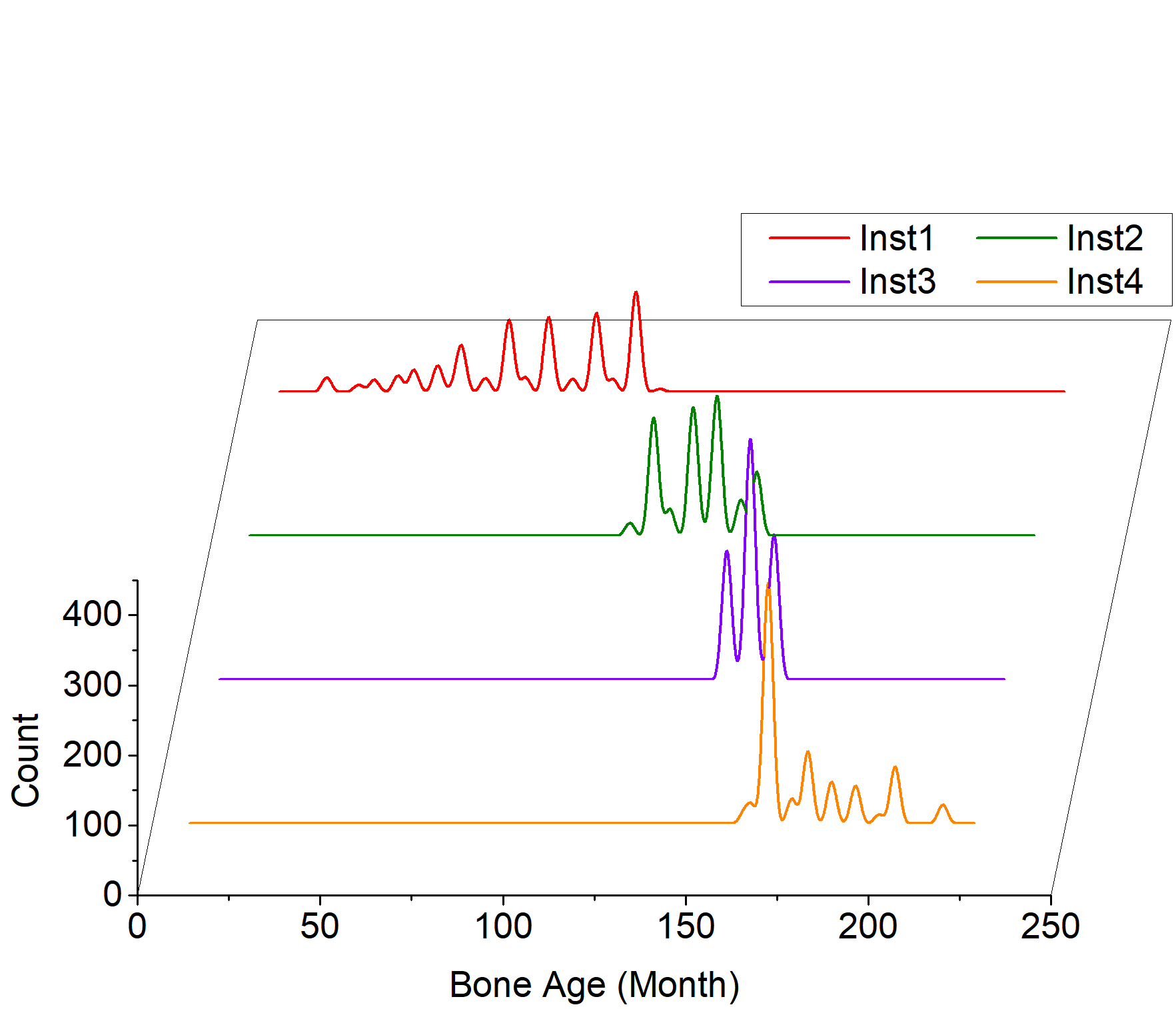}\\
 (a) Split 1, KS-0.02 & (b) Split 2, KS-0.63 & (c) Split 3, KS-0.97
 		\end{tabular}
	\end{center}
\vspace{-2mm}
	\caption{Three sets of simulated data partitions on BoneAge dataset \citep{halabi2019rsna}. The histogram of labels for each institution is shown. }
	\label{fig:BoneAge_Split}
\vspace{-2mm}
\end{figure*}

\textbf{Data Partitions.} We simulated three sets of data partitions on both Retina dataset \citep{KaggleRetina} and BoneAge dataset \citep{halabi2019rsna}, i.e., IID data partitions, non-IID data partitions with mild degree of label distribution skewness, and non-IID data partitions with high degree of label distribution skewness. We simulated 4 local institutions in each set of experiments. The detailed data partitions are shown in Fig.~\ref{fig:Retina_Split} and Fig.~\ref{fig:BoneAge_Split}. We applied the mean Kolmogorov-Smirnov (KS) statistics between every two institutions to measure the degree of label distribution skewness. KS=0 means IID data partitions, while KS=1 indicates identically different label distributions across institutions.

\textbf{Comparison Methods.} We compared our CWT+Replay and FedReplay with the following baseline methods for collaborative learning: \begin{itemize}
\item FedAVG \citep{mcmahan2016communication}, one of the most popular parallelized methods for collaborative learning. 
    We set the fractions of institutions that were selected for computation in each round to 1, i.e., all the local institutions were involved in the computation in each round. In addition, we set the number of iterations trained on local institution to $\lfloor |D_k|/b\rfloor$ in each round, where $|D_k|$ is the quantity of training samples in local institution $k$ and $b$ is the local minibatch size.


\item CWT \citep{chang2018distributed}, a type of serial collaborative learning. 
    Following ~\citep{Niranjan2020}, we set the number of iterations trained on local institution $k$ to $\lfloor |D_k|/b\rfloor$ to prevent performance drops from sample size variability.

\item SplitNN \citep{vepakomma2018split}, can be considered as a serial collaborative learning. 
    We choose the peer-to-peer mode version in \citep{vepakomma2018split} for collaborative training, where one institution first starts one round of training, then cyclically sends the partial of network model weights to the next institution for another round of training until model convergence. In SplitNN, in addition to the model weights, the intermediate feature maps and gradients are also communicated between the institutions and the central server. Similar to SplitNN, our CWT+Replay and FedReplay also require transferring the intermediate feature maps (latent variables) between the institutions and the central server.

\item FedAVGM \citep{hsu2019measuring}, an optimization method for FedAVG \citep{mcmahan2016communication}, which applies momentum optimizer on the server \citep{hsu2019measuring} to improve its robustness to non-IID data. 

\item FedAVG+Share \citep{zhao2018federated}, an optimization method for FedAVG \citep{mcmahan2016communication}. FedAVG+Share aims to improve the performance of FedAVG on non-IID data by globally sharing a small amount of data between all the institutions. Following \citep{zhao2018federated}, we distributed 5\% globally shared data among each institution in our experiments.
\item FedProx ~\citep{li2020federated}, an optimization method for FedAVG  \citep{mcmahan2016communication}, which introduces a proximal term to the local objective to help stabilize the model training. We tune the penalty constant $\mu$ in the proximal term from the candidate set \{0.001,  0.01, 0.1, 1\} on Split-2 for both Retina and BoneAge datasets, and apply the same penalty constant for all the remaining data partitions. Specifically, $\mu$ is set to 0.001 for both Retina and BoneAge dataset.
\end{itemize}

\begin{figure}[h]
	\centering
	\includegraphics[width=0.95\linewidth]{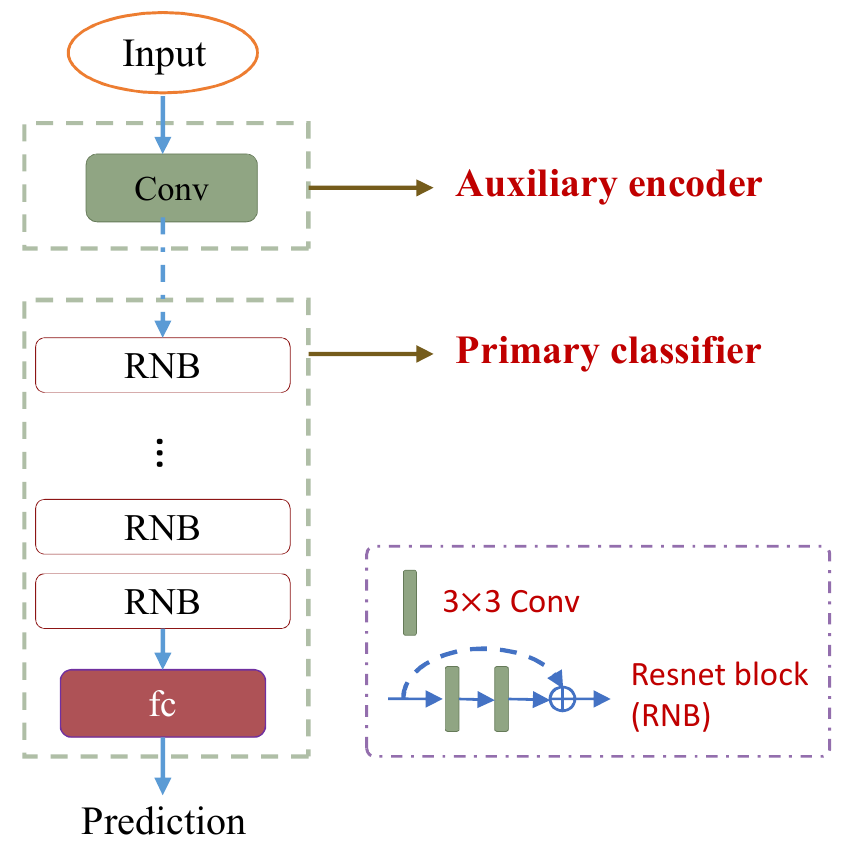}
	\vspace{-1.0mm}
	\caption{Detailed network architecture of FedReplay when ResNet34 \citep{he2016deep} is used as the backbone network. We omit ReLu, batch normalization and max pooling layers that follow each convolution layer. We split the ResNet34 \citep{he2016deep} into two parts, where the first part was used as the auxiliary encoder network, and the remaining part was used as the primary classifier network. }
	\label{fig:FedReplay_network}
	\vspace{-3.5mm}
\end{figure}

\begin{table}[t]
  \centering
  \caption{Catastrophic forgetting phenomenon of CWT \citep{chang2018distributed} on non-IID data partitions Split 3 of Retina dataset \citep{KaggleRetina}.  The prediction accuracy on 4 institutional train dataset from models achieved after one round of training on each institution is shown. In this experiment, the model was trained in cyclical order: Inst1$ \rightarrow$ Inst2 $\rightarrow$ Inst3 $\rightarrow$ Inst4 $\rightarrow$ Inst1 ...}
  \label{table:catastrophe_forgetting}
  \begin{tabular}{c|cccc}
    \hline
 \diagbox{Data}{Model} & Inst1 &   Inst2  &   Inst3  &    Inst4 \\
    \hline
Inst1 &  95.6\% &91.0\% & 69.9\% & 51.7\% \\
Inst2 & 83.4\% & 87.7\% & 69.0\% & 60.0\%  \\
Inst3 & 23.4\% &  53.7\% &  93.0\% & 91.0\% \\
 Inst4 & 9.4\% & 41.8\% & 95.7\% & 97.7\% \\
    \hline
  \end{tabular}
  \vspace{-0.5mm}
\end{table}
\textbf{Implementation Details.} All the methods described in this paper were implemented with Pytorch and optimized with SGD. We used the ResNet34 \citep{he2016deep} as the backbone network architecture for both the classification and regression tasks in all collaborative learning methods. For SplitNN \citep{vepakomma2018split}, we applied the data partitions with Split 2 as example, and run multiple experiments on both Retina \citep{KaggleRetina} and BoneAge dataset \citep{halabi2019rsna} to choose the optimized cut layer. Specifically, conv1 was applied as the cut layer since it provides the best performance in our experiments. We set batch size $b$ to 32, the learning rate $\eta$ to 0.001 and progressively decreased with scale 10 every 35 epochs. For auxiliary generator of CWT+Replay: we followed work in \citep{razavi2019generating} to set the hyper parameters for VQ-VAE-2. All the images in VQ-VAE-2 were pre-processed to resolution $256 \times 256$, and thus the latent layers of VQ-VAE-2 were with resolution $32 \times 32$ and $64 \times 64$ for top levels and bottom levels, respectively. Unlike in \citep{razavi2019generating}, we set the batch size to 32, and the learning rate $\eta_G$ to 0.0004 (halved every 150 epochs). Adam \citep{kingma2014adam} was applied as the optimizer.

We used ResNet34 \citep{he2016deep} as the backbone network for FedReplay. Shown in Fig.~\ref{fig:FedReplay_network}, we split the ResNet34 \citep{he2016deep} into two parts, where the first part (consisting of first convolutional, batch normalization, ReLU and max pooling layer) was used as the encoder network, while the remaining part was used as the primary classifier network. The training of the encoder network is achieved on one of the arbitrary institution with the institutional dataset and the desired task. We applied the whole ResNet34 \citep{he2016deep} to help train the encoder network. Specifically, we trained the ResNet34 \citep{he2016deep} with the attached institutional dataset on the learned task (classification task for Retina \citep{KaggleRetina} dataset, and regression task for BoneAge dataset \citep{halabi2019rsna}) until model convergence. The first part of the trained ResNet34 \citep{he2016deep} were then used as the auxiliary encoder network. The training parameters for the auxiliary encoder network and primary classifier were the same to the primary classifier model in CWT+Replay.

\subsection{Performance Evaluation}
\begin{figure*}[t]
\scriptsize
	\begin{center}
		\begin{tabular}{cc}
 \includegraphics[width=0.48\linewidth]{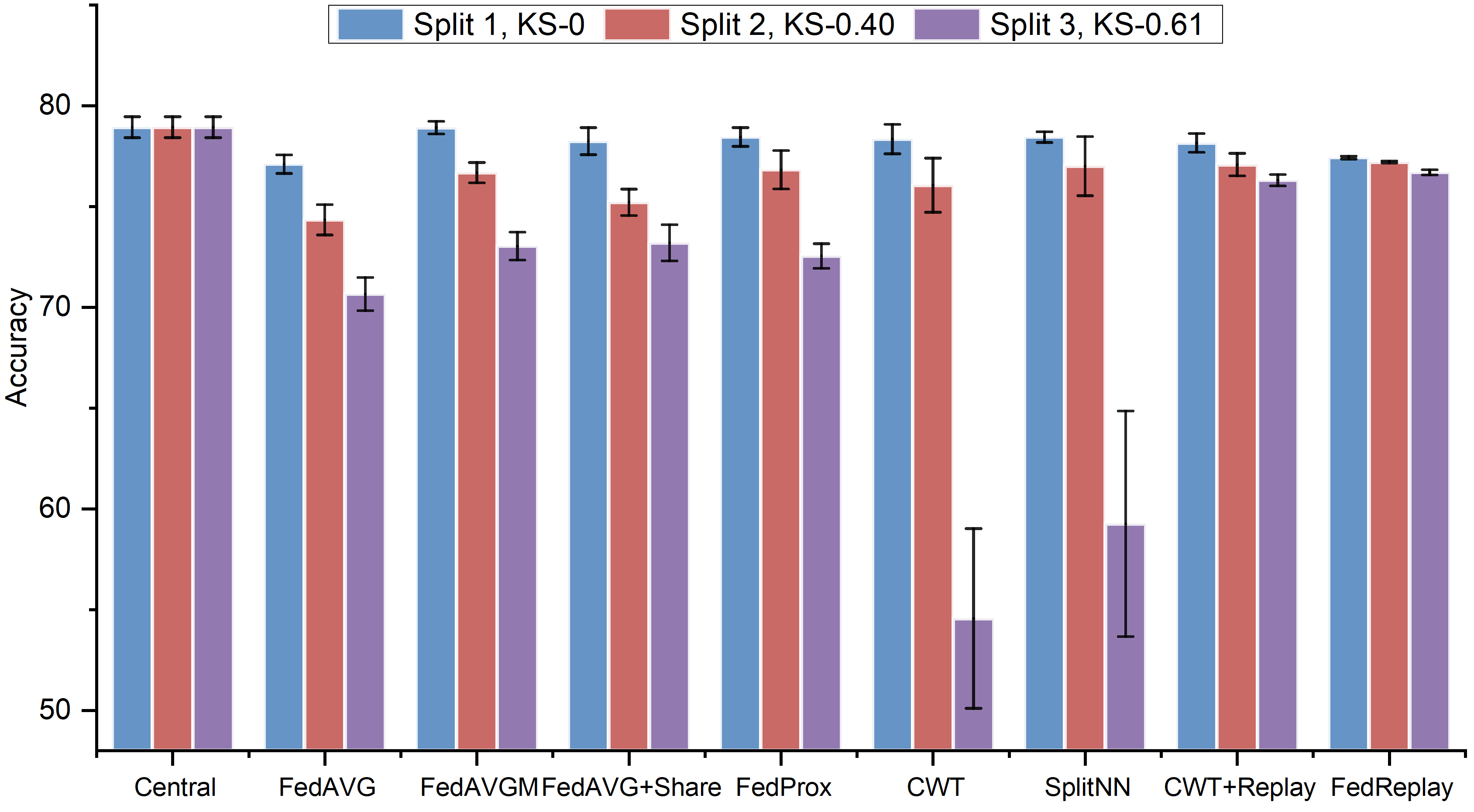} &
 \includegraphics[width=0.48\linewidth]{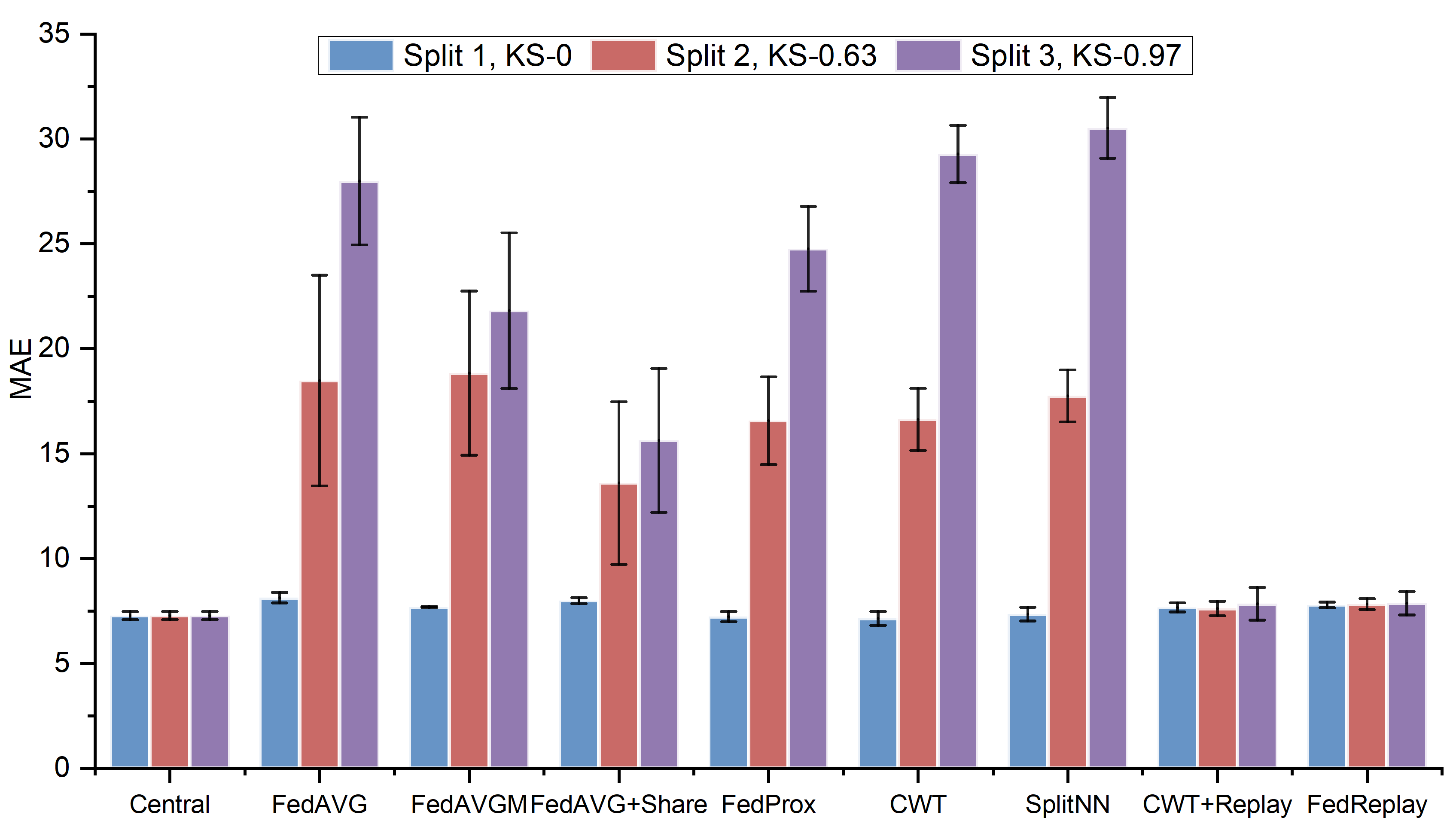}\\
 (a) Predicted accuracy on Retina dataset & (b) Predicted MAE on BoneAge dataset
 		\end{tabular}
	\end{center}
\vspace{-3mm}
	\caption{Performance evaluation on Retina dataset \citep{KaggleRetina} and BoneAge dataset \citep{halabi2019rsna}.
Mean and standard deviation test accuracies (the higher the better) and MAE loss (the lower the better) were obtained with 4 runs. We use the
same setting for the following experiments.}
	\label{fig:performance}
\vspace{-3.5mm}
\end{figure*}

In this section, we evaluated the performance of different collaborative learning methods, assessing prediction accuracy for Retina dataset \citep{KaggleRetina}, and mean absolute error (MAE) loss for BoneAge dataset \citep{halabi2019rsna}. The performance on the centrally hosted data was applied as the benchmark performance.

As shown in Fig.~\ref{fig:performance}, all the compared collaborative learning methods show comparable performance to the benchmark centrally hosting method in the case where data at different institutions is homogenous (Split 1). However, the performance of the standard CWT \citep{chang2018distributed}, SplitNN \citep{vepakomma2018split}, and FedAVG \citep{mcmahan2016communication} drop significantly with the increasing degree of label distribution skewness in data across institutions. For example, the MAE loss of CWT \citep{chang2018distributed}, SplitNN \citep{vepakomma2018split}, and FedAVG \citep{mcmahan2016communication} on BonAge dataset  \citep{halabi2019rsna} increase from $6.88 \pm 0.04$, $6.80 \pm 0.05$, $7.26 \pm 0.07$ on data Split 1 to $27.88 \pm 2.82$, $30.28 \pm 1.47$, $30.87 \pm 0.07$ on data Split 3, respectively.

The serial collaborative learning methods, such as CWT \citep{chang2018distributed} and SplitNN \citep{vepakomma2018split}, always suffer from catastrophic forgetting when heterogeneity exists in data distributions across institutions. The model tends to abruptly forget what was previously learned information when it transfers to next institution. For example, in the Split 3 experiment of Retina dataset \citep{KaggleRetina}, when transferring the model trained in Inst2 to the next Inst3, the model learned in Inst 3 tends to forget what was previously learned information from Inst2, and the prediction accuracy on Inst2 is dropped from original 87.7\% to 69.0\% (see the second row in Table~\ref{table:catastrophe_forgetting}). While for parallelized methods, such as FedAVG \citep{mcmahan2016communication}, the data used for training at each local
institution is sampled from an institution-specific distribution, which is a biased estimator of the central target distribution
if heterogeneity exists. The learned weights in different institutions will diverge severely when high skewness exists in data across institutions, and
the synchronized averaged central model will lose accuracy or even completely diverge \citep{zhao2018federated}. The existing optimized methods, such as FedAVGM \citep{hsu2019measuring}, FedProx \citep{li2020federated} and FedAVG+Share \citep{zhao2018federated} may help alleviate the model divergence problem, but they still suffer from performance drops in highly skewed non-IID data partitions, such as the MAE loss $21.81 \pm 3.71$ and $15.63 \pm 3.43$  of FedAVGM \citep{hsu2019measuring} and FedAVG+Share \citep{zhao2018federated} on Split 3, compared to the MAE loss $7.49 \pm 0.032$, and $7.99 \pm 0.14$ on homogenous Split 1.
As a comparison, our generative replay strategy helps generate synthetic images closely resembling the studied
participants (CWT+Replay) or provide accessible to the union of the latent variables (FedReplay), thus can avoid the performance drops even on highly skewed heterogeneity cases.

\subsection{Application to a Real-World Federated Dataset}
We further evaluated our method on a real-world federated dataset, International Brain Tumor Segmentation (BraTS) 2017 challenge \citep{menze2014multimodal,bakas2017advancing}, and compared it to the state-of-the-art collaborative learning methods CWT \citep{chang2018distributed}, FedAVG \citep{mcmahan2016communication}, and FedAVG+Share \citep{zhao2018federated}. The performance on the centrally hosted data was applied as the benchmark performance.

\begin{figure}[h]
\scriptsize
	\begin{center}
		\begin{tabular}{c}
 \includegraphics[width=0.98\linewidth]{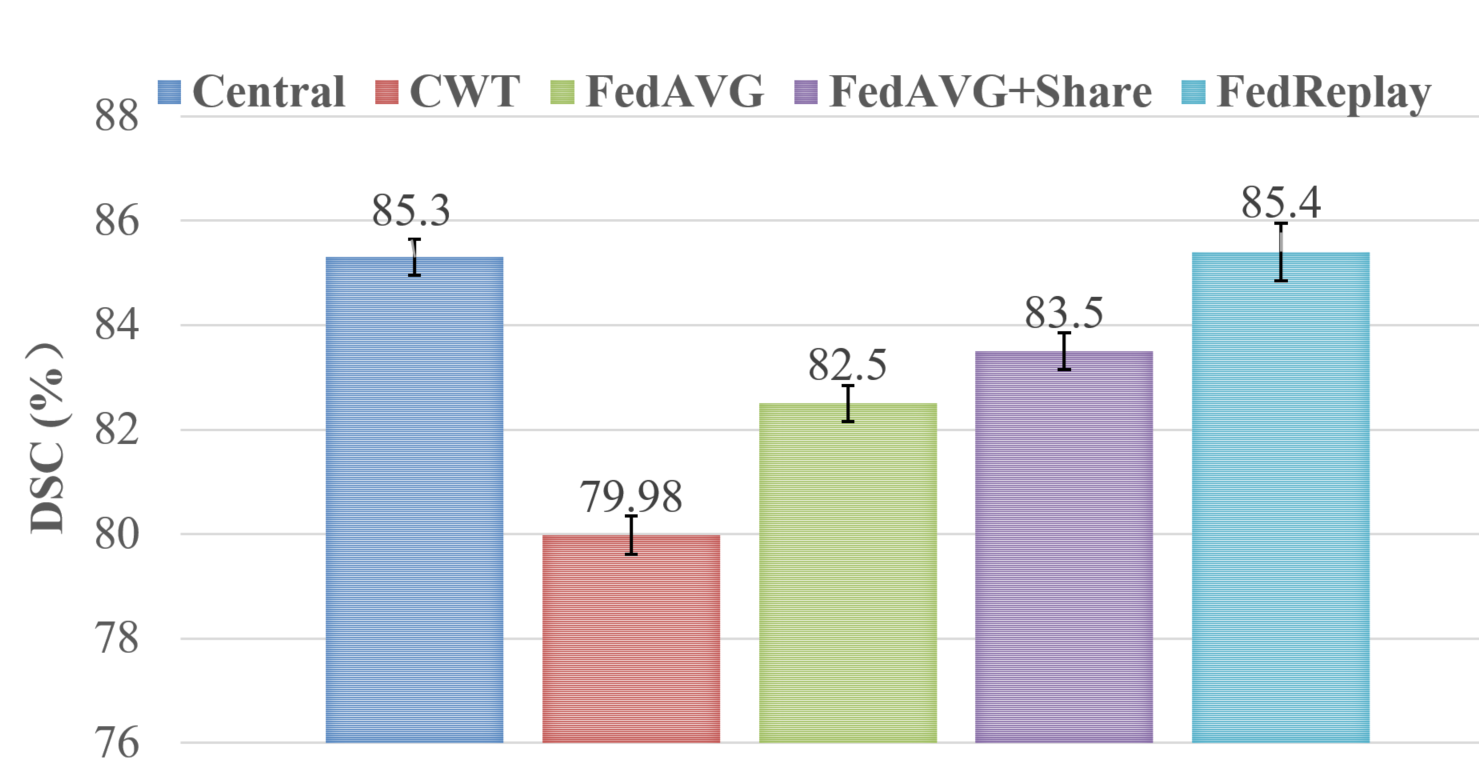}
 		\end{tabular}
	\end{center}
\vspace{-6mm}
	\caption{Mean predicted dice similarity coefficient (DSC) over 10 institutions trained with CWT, FedAVG, FedAVG+Share, and the proposed FedReplay on BraTs 2017, as a comparison to the benchmark method (Central) on the centrally hosted data. Our FedReplay outperforms all the competing collaborative learning methods on the real-world federated dataset.}
	\label{fig:dsc_brats}
\end{figure}

BraTs 2017 is a multi-institutional pre-operative multimodal MRI scans of glioblastoma (GBM/HGG) and lower grade glioma (LGG). Each subject consists of four modal MRI scans: a) native (T1), b) post-contrast T1-weighted, c) T2-weighted, and d) T2 Fluid Attenuated Inversion Recovery (T2-FLAIR) volumes, and an associated voxel-level annotation of ``tumor core'', ``enhancing tumor'', and ``whole tumor''. We used HGG patients in our study, applied the T2-FLAIR modality as the input, and evaluated on the binary whole tumor segmentation task. We used the training dataset of BraTs 2017 (consists of 210 subjects collected from 10 different institutions \footnote{We thank Spyridon Bakas, one of the authors of BraTs challenge, for providing real institutional details of BraTs 2017.}) to form our institutional training and test datasets. Specifically, we random selected 45 subjects as the test dataset, used the remaining 165 subjects for the training dataset. We run with three trials and applied the Dice Similarity Coefficient (DSC) \citep{dice1945measures} to measure similarity between the predicted results and the ground truth mask label.

We used U-net \citep{ronneberger2015u} pre-trained on brain MRI segmentation dataset \citep{buda2019association} as the baseline network. U-net consists of four levels of blocks and a bottleneck layer with 512 convolutional filters. Each block contains an encoding part with two convolutional layers (followed by batch normalization and ReLU activation function) and one max pooling layer,  and a decoding part with two convolutional layers (followed by batch normalization and ReLU activation function) and an up-convolutional layer. Skip connections are applied between each encoding layer and its corresponding layer in the decoding part. For FedReplay, we split the U-net at the first encoding block, applied the first encoding as the auxiliary encoder network, used the remaining part as the primary classifier. We trained the auxiliary encoder on a institution that contains the largest number of subjects, and then distributed the auxiliary encoder to all the rest institutions for feature extraction. We extracted three consecutive axial slices as the input, all the images were resized to $256\times256$ and then cropped to $224\times224$. All our experiments used a batch size of 32 and learning rate 1e-4 with Adam optimizer. All the models were trained with total 35 communication rounds.

We report mean DSC over 10 institutions trained with CWT, FedAVG, FedAVG+Share, and the proposed FedReplay on BraTs 2017 in Figure~\ref{fig:dsc_brats}.
FedReplay outperforms all the competing collaborative learning methods on the real-world federated dataset, achieves comparable performance to the benchmark method (Central) on the centrally hosted data, i.e., 85.4\% of our FedReplay as a comparison to 85.3\% of Central.

\subsection{Privacy Analysis of Data Leakage}\label{section:exp_privacy}
We evaluate the privacy protection capability of FedReplay by comparing the quality of the recovered images from FedAVG and FedReplay according to Eq.~\ref{FedAVG_DL} and Eq.~\ref{FedReplay_DL}. We apply the Peak Signal-to-Noise Ratio (PSNR), a ratio between the maximum value (power) of an image and the mean squared error between the target image and recovered image. We measure the PSNR of the reconstruction of $224 \times 224$
Retina images over a random selected 50 images of the test dataset. The higher the PSNR value, the better the reconstructed image quality. We apply AdamW to optimization Eq.~\ref{FedAVG_DL} and Eq.~\ref{FedReplay_DL}, and all optimization runs up to 20,000 iterations.

We use ResNet34 as the baseline network. ResNet34 consists of several sequential Residual blocks and can be listed as \{Conv1, Layer1, Layer2, Layer3, Layer4, FC\}. We evaluate three sets of auxiliary encoder network (extracted from ResNet34) for FedReplay: 1) Conv1: a Convolutional layer following up a ReLu, a batch normalization and a max pooling layer, 2) Layer2: Conv1 with two residual blocks, and 3) Layer 3: Conv1 with three Residual blocks.

\begin{table*}[t]
  \centering
  \caption{Comparison of FedReplay and FedAVG in terms of prediction accuracy and average restoration performance on Split-3 of Retina datasets. FedReplay achieves better prediction accuracy on heterogenous data partitions while at the same time shows similar or even better patient privacy capability (worse reconstruction image quality) than FedAVG. }
  \label{table:acc_restoration}
  \begin{tabular}{c|cccc}
    \hline
  & FedAVG &   Ours (Conv1)& Ours (Layer 2) & Ours (Layer 3)  \\
    \hline
ACC &  70.65 $\pm $ 0.83 & \textbf{77.20 $\pm $ 0.13} & 76.48 $\pm $ 0.63 &  75.00 $\pm $ 0.98   \\
PSNR &  \textbf{11.82 $\pm $ 1.86} &  11.64 $\pm$  0.98 & 10.48 $\pm$ 1.23 & 9.22 $\pm$  1.28   \\
    \hline
  \end{tabular}
\end{table*}

\begin{figure*}\footnotesize
	\centering
	\begin{center}
		\begin{tabular}{ccccc}

GT. & FedAVG & Ours (Conv1) & Ours (Layer2) & Ours (Layer3)\\

	\includegraphics[width=0.192\linewidth]{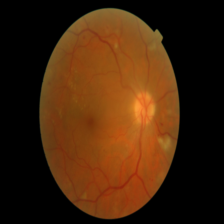}& \hspace{-4mm}
	\includegraphics[width=0.192\linewidth]{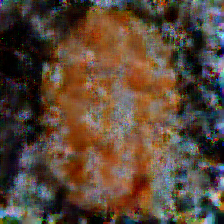}&  \hspace{-4mm}
	\includegraphics[width=0.192\linewidth]{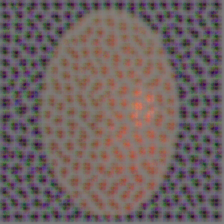} &  \hspace{-4mm}
	\includegraphics[width=0.192\linewidth]{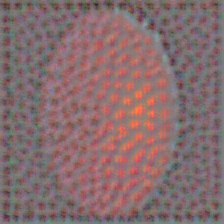}&  \hspace{-4mm}
	\includegraphics[width=0.192\linewidth]{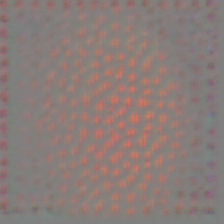}\\
& 12.75 db & 10.38 db & 8.97 db & 7.97 db \\
	\includegraphics[width=0.192\linewidth]{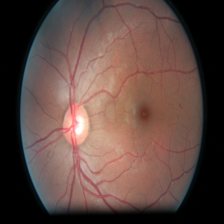}& \hspace{-4mm}
	\includegraphics[width=0.192\linewidth]{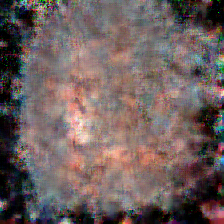}&  \hspace{-4mm}
	\includegraphics[width=0.192\linewidth]{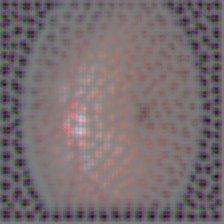} &  \hspace{-4mm}
	\includegraphics[width=0.192\linewidth]{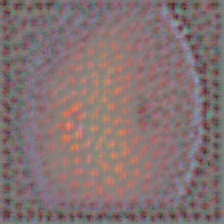}&  \hspace{-4mm}
	\includegraphics[width=0.192\linewidth]{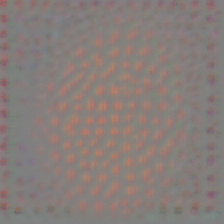}\\
& 14.75 db & 13.05 db & 11.74 db & 10.61 db \\

		\end{tabular}
	\end{center}
	\vspace{-3mm}
	\caption{Reconstruction of two images from FedAVG with deep gradient inversions and our FedReplay with model inversion attacks. We experiment with three neural network architectures for the auxiliary encoder of FedReplay, from simple Conv1 layer, to several Residual blocks. FedReplay shows similar privacy protection capability to the baseline FedAVG when shallow network is applied as the encoder network, and superior protection capability when deeper neural network is applied (Ours Layer3).}
	\label{fig:reconstructed_imgs}
	\vspace{-3mm}
\end{figure*}

Table~\ref{table:acc_restoration} compares the prediction accuracy and the privacy protection capability of FedAVG and the proposed FedReplay on data partition Split-3 of Retina dataset. It is not easy to reconstruct a recognizable image from both FedAVG and FedReplay under current attacker technique when the input image has high resolution($224\times224$), as the low PSNR shown in Table~\ref{table:acc_restoration} and no clear patterns shown in Fig.~\ref{fig:reconstructed_imgs}. In addition, the flexibility design of the dual model architecture of our FedReplay allows the room for better privacy protection capability by applying a deeper network as the auxiliary encoder while only slightly impeding the final prediction performance. See the nearly unrecognizable results from Layer3 shown in Fig. 7 and the decent prediction performance 75.00\% shown in Table 3 (compared to 70.65\% of FedAVG). 
\section{Conclusion}
In this paper, we present a novel generative replay strategy to address the challenge of data heterogeneity across institutions in collaborative learning. Our generative replay strategy is flexible to use. It can either be incorporated into existing collaborative
learning methods to improve their capability of handling data heterogeneity across institutions, or be used as
a novel and separate collaborative learning framework (FedReplay) to reduce communication cost and mitigate the privacy cost. Unlike existing methods that either design sophisticated ways to control the optimization strategy or share partial global data to mitigate the performance drops from data heterogeneity across institutions, our generative replay strategy is a flexible and straightforward alternative solution, which provides new insight for the development of collaborative learning methods in real applications.

One concern of our generative replay technique is that the task performance is heavily dependent on the quality of the generator when it is applied to existing collaborative learning methods. However, this is also largely alleviated by the progress of training generative models. In addition, our FedReplay relaxes this constraint by training a primary model based on the extracted latent variables, and thus do not need to train a high quality generator for synthesizing images that closely resembles the input images. As a contrast, the training of the auxiliary encoder network for FedReplay can be easily achieved on one of the arbitrary institutions with the learned task.


\section*{Acknowledgment}
This work was supported in part by a grant from the NCI, U01CA242879.

\section*{CRediT authorship contribution statement}
Liangqiong Qu: Conceptualization, Investigation, Formal analysis, Methodology, Software, Validation, Writing – original draft,
Writing – review \& editing. Niranjan Balachandar: Investigation,
Validation, Writing – review \& editing. Miao Zhang: Investigation,
Software. Daniel Rubin: Conceptualization, Formal analysis, Fund-
ing acquisition, Writing – review \& editing.

\end{document}